\documentclass{ecai} 
\usepackage{latexsym}
\usepackage{amssymb}
\usepackage{amsmath}
\usepackage{amsthm}
\usepackage{booktabs}
\usepackage{enumitem}
\usepackage{graphicx}
\usepackage[font=small]{caption}
\usepackage{color}
\usepackage[ruled,vlined,linesnumbered]{algorithm2e}
\usepackage{algpseudocode}
\usepackage{newfloat}
\usepackage{listings}
\usepackage{float} 
\usepackage{graphicx}  
\usepackage{float}   
\usepackage[hidelinks]{hyperref}

\newcommand{\BibTeX}{B\kern-.05em{\sc i\kern-.025em b}\kern-.08em\TeX}

\begin{document}

\begin{frontmatter}
\paperid{3531} 
\title{A Step Towards Robust Unsupervised Domain Adaptation via Fine-Tuning and Reinforcement Learning}

\author[A]{\fnms{Sushant Dagaji}~\snm{Desale}}
\author[A]{\fnms{Rahul}~\snm{Mishra}\thanks{Corresponding Author. Email: rahul\_mishra@iitp.ac.in},}
\author[A]{\fnms{Ashutosh Kumar}~\snm{Sinha}} 

\address[A]{Indian Institute of Technology Patna}

\begin{abstract}
Adversarial robustness in Unsupervised Domain Adaptation (UDA) remains a significant challenge due to noisy pseudo labels and inherent distributional shifts between the clean source and adversarially perturbed target domains. Existing approaches often fail to achieve an optimal trade-off between robustness and accuracy, as pseudo-labels generated by domain-adapted models tend to introduce classification errors under adversarial attacks. In this work, we propose \textbf{SFT+RL}, a two-stage robust UDA framework that integrates Supervised Fine Tuning (SFT) and Reinforcement Learning (RL) on top of CLIP's pre-trained visual encoder. In the SFT stage, we adversarially fine-tune a linear classifier using PGD-based perturbations over the labelled source domain while partially unfreezing CLIP's projection layer. It allows adaptation to adversarial noise while preserving CLIP's rich semantic priors. We introduce a confidence-guided pseudo-labeling strategy in the RL stage to annotate unlabeled target samples progressively. Pseudo labels are filtered using a decaying confidence threshold to balance quality and coverage, and the model is trained on a composite dataset formed by combining clean source samples with high-confidence target samples. Adversarial training is applied to mixed batches of clean and adversarial examples to enhance cross-domain robustness. Comprehensive evaluations on three benchmark datasets OfficeHome~\cite{tomm-ude}, PACS~\cite{pacs}, and VisDA~\cite{visda} demonstrate the effectiveness of our approach. Notably, \textbf{SFT+RL} achieves average improvements of \textbf{10.2\%} in clean accuracy and \textbf{15.8\%} in adversarial robustness across all three datasets, outperforming existing state-of-the-art methods.

\end{abstract}

\end{frontmatter}

\section{Introduction}
Vision Language Models (VLMs) such as Contrastive Language-Image Pretraining (CLIP)~\cite{radford2021learning} have greatly improved advanced zero-shot image classification by learning joint visual, textual embedding spaces via large-scale contrastive pretraining. Such models project images and text onto a shared semantic space, enabling the direct matching of novel visual concepts to their natural language descriptions without task-specific fine-tuning~\cite{radford2021learning}. Despite their impressive generalization of clean data, VLMs exhibit colossal failures when confronted with adversarial perturbations. It reveals the fundamental challenges that undermine VLMs' robustness in real-world settings.

\begin{figure}[t]
  \centering
  \includegraphics[width=1.05\linewidth]{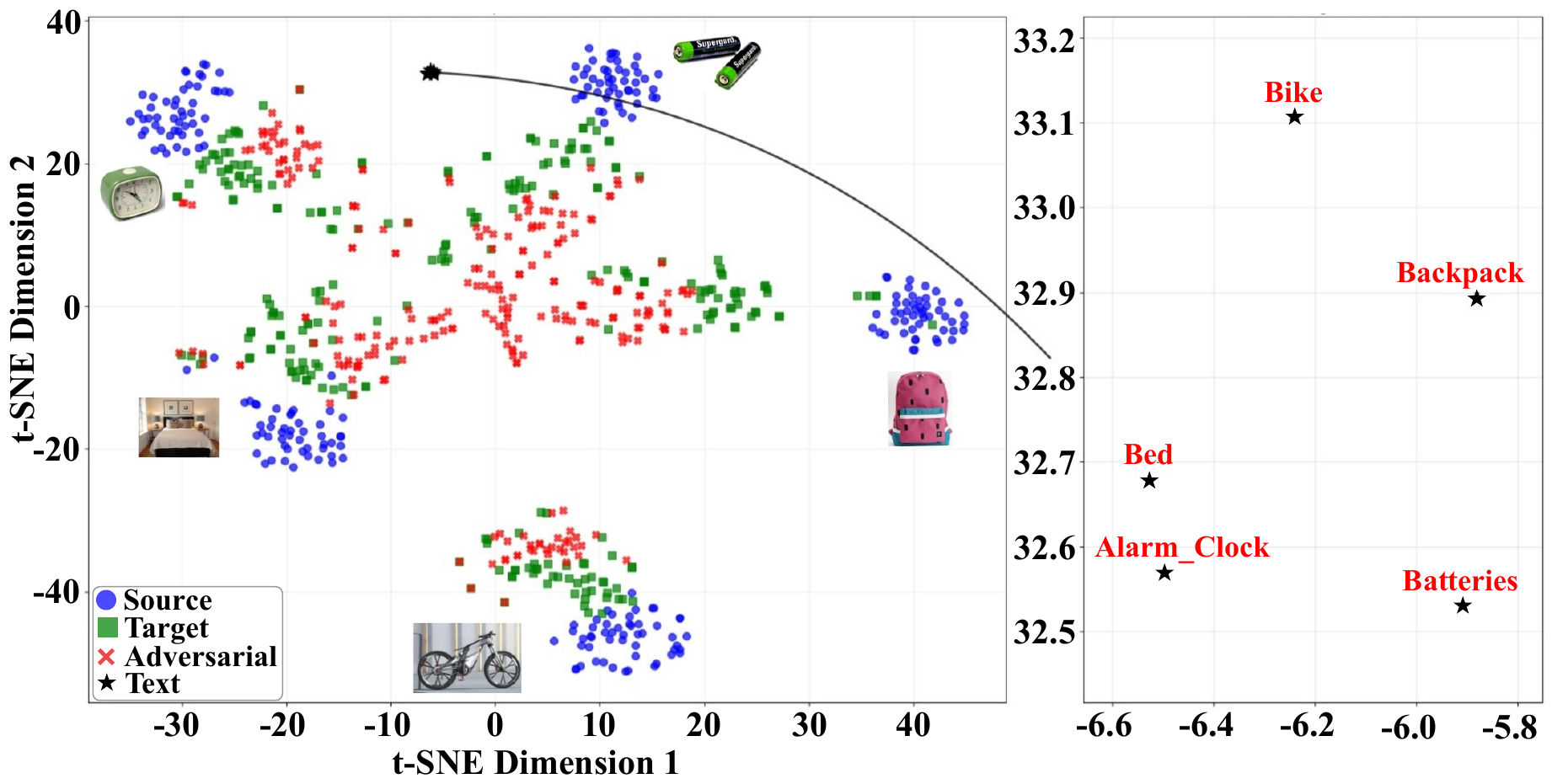}
  \caption{Adversarial domain alignment challenge: Clean source samples (\textcolor{blue}{blue}), clean target samples (\textcolor{green}{green}), and adversarial target samples (\textcolor{red}{red}) tend to form indistinguishable clusters in CLIP’s embedding space.\vspace{0.5cm}}
  \label{fig:problem_statement}
\end{figure}

CLIP’s visual encoder yields transferable feature representations across many natural domains. However, the adversarial attacks induce non-systematic shifts outside the distributional assumptions of domain adaptation~\cite{wang2018deep}. Projected Gradient Descent (PGD) style attacks~\cite{madry2018towards} can mostly change the visual embedding using small $\ell_{\infty}$ bounded perturbations while keeping the corresponding text embedding mostly unaltered, decoupling the two modalities~\cite{liang2022enhancing}. Such decoupling creates ambiguous regions in feature space where clean and perturbed target sample clusters are indiscriminate, as shown in Figure~\ref{fig:problem_statement}. It leads to erroneous cross-modal retrieval and classification even when original text prompts remain semantically separated.

CLIP's text encoder often lacks the fine-grained discriminative capacity to distinguish adversarially perturbed images~\cite{zhou2022robust}. Over-parameterized text encoders activate spurious, non-semantic dimensions under adversarial noise, a phenomenon we refer to as \emph{textual redundancy}~\cite{tanwisuth2023textual}. Simultaneously, source-trained visual projections tend to overfit class-agnostic texture and color cues. Adversarial attacks can utilize such overfitting to induce misclassification, an effect we term \emph{visual overfitting}~\cite{salman2021adversarial}. These complementary failure modes exacerbate cross-modal misalignment, as the text encoder cannot compensate for the corrupted visual features.

Prior work in adaptation and robustness suffers from various incompatibilities with VLMs. Source dependent methods~\cite{saito2019semi, kang2019contrastive, na2021wilds} rely on abundant labelled source data and complete network fine-tuning, diminishing CLIP’s promise of zero-shot transfer. Unsupervised domain adaptation techniques~\cite{wang2020transferable, yang2022domain, sharma2021generalize} lack explicit adversarial defences and fail when target inputs are maliciously perturbed. Prompt tuning approaches~\cite{zhou2022robust, gao2021making} can bolster clean-data performance but degrade sharply under attack (as shown in Figure~\ref{fig:problem_statement}), while recent vision language adapters~\cite{tanwisuth2023textual} neglect to preserve cross-modal alignment during adaptation, leading to fragmented embeddings. These challenges in the prior studies have created the need for a method that simultaneously addresses \textit{adversarial robustness, domain shift, and modality alignment in VLMs}.

This work presents, \textbf{SFT+RL}, a two-phase adaptation framework that seamlessly combines adversarial fine-tuning with a confidence-driven cross-modal curriculum to solve the challenges of CLIP in adversarial and domain-shifted settings. In the first phase, \textbf{Supervised Fine Tuning (SFT)}, we anchor the model's visual representations by incorporating on-the-fly PGD perturbations into the source domain training loop. Next, the second stage uses \textbf{Reinforcement Learning (RL)} on unlabeled target data to improve pseudo labels under adversarial scrutiny, building upon the SFT's robust initialization. 

STF preserves pre-learned semantic priors and mitigates overfitting to source textures by partially unfreezing only CLIP's final projection layer while keeping the bulk of the ViT-B/32 backbone frozen~\cite{radford2021learning}. Such a targeted unfreezing schedule allows the classifier to adapt rapidly to adversarial noise without compromising the model's zero-shot generalization capabilities. Further, RL assigns each target sample to the class with maximum posterior probability and retains only those predictions exceeding a confidence threshold $\tau$. The threshold functions act as a curriculum to balance label quality against sample coverage. It starts from a high confidence regime and decays each cycle gradually but never below a minimum floor.
Specifically, this process alternates between updating the text encoder using adversarially perturbed visuals and adapting the projection layer through refined pseudo labels, which enforces cross-modal consistency even under attack.

Our framework distinguishes itself from existing methods in four key aspects. First, unlike source-dependent approaches such as TENT~\cite{gao2021making} or SHOT~\cite{liang2020we}, SFT+RL requires only unlabeled target inputs at test time, preserving CLIP’s zero-shot deployment paradigm. Second, it is intrinsically adversarially robust, unlike fully unsupervised UDA techniques that omit explicit defence mechanisms~\cite{tanwisuth2023textual}. Third, by alternating updates between visual and textual components, our method explicitly maintains cross-modal alignment, a feature lacking in prior vision language adapters. Finally, SFT+RL is fully compatible with CLIP’s architecture, leveraging its pre-trained weights without necessitating extensive network rewrites.

\noindent $\bullet$ The contributions and novelties of STF-RL are as follows:
\vspace{-0.2cm}
\begin{itemize}
    \item \textbf{Robust CLIP Adaptation:} We introduce SFT+RL, a two-phase framework that unifies supervised adversarial fine-tuning on source data with reinforcement-style self-training on unlabeled target samples. The framework blends PGD-hardened source training with confidence-driven pseudo-label refinement to enhance adversarial robustness of CLIP.

    \item \textbf{Anchoring with Partial Unfreezing:} Next, we propose a layer-wise unfreezing schedule to preserve CLIP’s powerful pre-trained representations while adapting to adversarial noise. Initially, we only fine-tuned CLIP’s projection layers on PGD perturbed source examples to harden the model against attacks. Gradually, additional projection parameters are unfrozen to ensure the ViT-B/32 backbone retains its zero-shot generalization ability.

    \item \textbf{Cross Modal Curriculum:} We further develop a pseudo-label refinement mechanism that dynamically decays the confidence threshold across training cycles. The classifier and text encoder are first updated using high-confidence target samples under adversarial images. Then, as the model becomes more confident, lower-confidence instances are gradually added. This dual-encoder curriculum enforces robust alignment between visual features and text embeddings, even in perturbations.

    \item \textbf{Experiments:} Finally, in the experimental evaluation under standard 20-step PGD attacks ($\epsilon = 2/255$),  SFT+RL sets a new performance bar on three challenging benchmarks. We achieve improvement in robustness and accuracy across OfficeHome~\cite{tomm-ude}, PACS~\cite{pacs}, and VisDA~\cite{visda} compared to state-of-the-art methods such as DART~\cite{wang2024dart}, SRoUDA~\cite{zhu2023srouda}, ARTUDA~\cite{yang2021exploring}, and adversarially-robust UDA variants including UDA+AT~\cite{madry2017towards}, UDA+TRADES~\cite{zhang2019theoretically}, and UDA+MART~\cite{madry2017towards}. Specifically, we observe average robustness gain at $\epsilon = 2/255$ of 34.5\%, 7.5\%, and 5.4\% on the OfficeHome, VisDA, and PACS datasets, respectively. 

\end{itemize}





\begin{figure*}[htbp]
\centering
\small
\includegraphics[width=14cm,height=7cm]{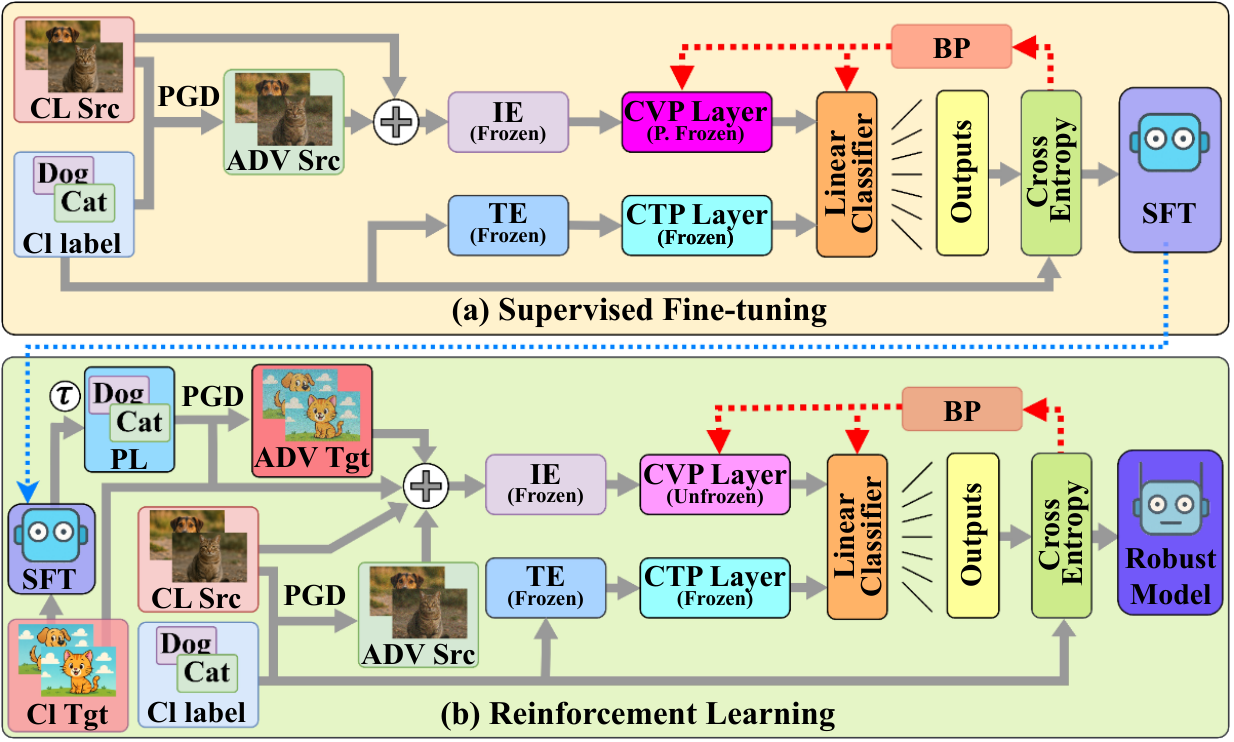}
\vspace{0.1cm}
\caption{Schematic representation of the SFT+RL framework: (a) In this stage, the model is fine-tuned by updating only the CLIP Vision Projection (CVP) layer (partially trainable) and the Linear Classifier (fully trainable), while keeping the Image Encoder (IE), Text Encoder (TE) and CLIP Text Projection (CTP) layer completely frozen. (b) This stage builds on the fine-tuned model to perform adversarial training on the target domain. It utilizes Clean Labels from the source domain and Pseudo Labels (PL) from the target domain. During this stage, both the CVP and Linear Classifier are fully trainable, while the IE, TE and CTP remain frozen. (BP: Backpropagation, SFT: Supervised Fine Tuning, CL: Clean, PGD: Projected Gradient Descent, ADV: Adversarial)
}       
\vspace{0.5cm}
\label{master}
\end{figure*}

\section{Preliminaries}

\subsection{Scenario and Problem Setting}
We consider the problem of adversarially robust UDA, where the objective is to learn a robust classifier that generalizes well across domains in the presence of adversarial perturbations. We consider the labeled source domain dataset denoted by $\mathcal{D}_s = \{(x_i^s, y_i^s)\}_{i=1}^{n}$, where $x_i^s \in \mathbb{R}^{H \times W \times 3}$ represents input images and $y_i^s \in \{1, \ldots, C\}$ denotes the corresponding class labels across $C$ classes. Additionally, we consider an unlabeled target domain dataset $\mathcal{D}_t = \{x_j^t\}_{j=1}^{m}$, where the underlying label distribution differs from the source domain due to a domain shift.

The primary challenge lies in learning a hypothesis $h(x) = g(f_{\text{CLIP}}(x))$, where $f_{\text{CLIP}}$ is a frozen visual encoder from the CLIP (ViT-B/32) model that provides rich semantic representations, and $g$ is a trainable linear classifier head. The goal is to ensure that $h(x)$ performs accurately on clean and adversarially perturbed target samples despite having no access to target labels during training.

In doing the above step, a key difficulty arises in aligning the feature distributions between the clean source $\mathcal{D}_s$ and adversarially perturbed target domain $\mathcal{D}_t^{\text{adv}}$, as well as ensuring consistency between pseudo-labeled target samples $\mathcal{D}_t^{\text{pseudo}}$ and their adversarial counterparts. It introduces the need for robust feature alignment and reliable pseudo-labeling under adversarial conditions.

\subsection{Adversarial Training}
Adversarial training is a defense mechanism where models are optimized on both clean and adversarially perturbed samples. Perturbations are typically generated using iterative attacks such as Projected Gradient Descent (PGD). Given an input $x$ and label $y$, the adversarial example $\hat{x}$ is obtained via iterative updates:
\begin{equation}
\hat{x}^{k+1} = \text{Clip}_\epsilon \left( \hat{x}^k + \alpha \cdot \text{sign}\big(\nabla_{\hat{x}^k} \mathcal{L}(h(\hat{x}^k), y)\big) \right),
\end{equation}
where $\epsilon$ bounds the perturbation, $\alpha$ is the step size, and $\mathcal{L}$ denotes the classification loss. Training on such adversarial examples improves robustness by making the model resistant to worst-case perturbations.



\section{Methodology}
Our proposed \textbf{SFT+RL framework} combines adversarial robustness with domain adaptation through a two-stage training process: 
\begin{enumerate}
    \item \textbf{Supervised Fine-Tuning (SFT)} on the source domain with adversarial training
    \item \textbf{Reinforcement Learning (RL)} on pseudo-labeled target data to adapt to the target domain.
\end{enumerate}
SFT+RL leverages CLIP’s visual encoder and trains a classifier head to align cross-domain features for adversarial robustness. Figure~\ref{master} illustrates the training pipeline of the proposed framework, which Algorithm~\ref{alg:rcaa} summarizes all the steps.

\subsection{Framework Architecture}
Our proposed framework architecture builds upon the powerful representational capabilities of CLIP's visual encoder while introducing a lightweight trainable classifier head to facilitate effective domain adaptation and adversarial robustness. The architecture comprises two main components, as depicted in Figure~\ref{master}:

\begin{enumerate}[label=\alph*)]
    \item \textbf{Frozen CLIP Image Encoder:} 
    We utilize the pre-trained image encoder from the CLIP (Contrastive Language-Image Pretraining) model, denoted as $f_{\text{CLIP}}(\cdot)$, which projects input images $\mathbf{x} \in \mathbb{R}^{H \times W \times C}$ into a $512$-dimensional embedding space. This encoder remains frozen in training to preserve the generalizable visual features learned from large-scale contrastive pretraining.
    
    \item \textbf{Trainable Linear Classifier:} 
    On top of the frozen encoder, we add a lightweight linear classifier $g(\cdot)$ that maps the extracted CLIP features to the class logits. The overall model prediction function is as follows:
    \begin{equation}
        h(\mathbf{x}) = g(f_{\text{CLIP}}(\mathbf{x})),
    \end{equation}
    where $h(\mathbf{x}) \in \mathbb{R}^K$ is the predicted logits for $K$ target classes.
\end{enumerate}

\SetAlFnt{\small}
\begin{algorithm}[h]
\caption{\textbf{SFT+RF Algorithm.}}
\label{alg:rcaa}
\KwIn{Source $\mathcal{D}_s=\{(x_i^s,y_i^s)\}_{i=1}^n$, Target $\mathcal{D}_t=\{x_j^t\}_{j=1}^m$, Pre-trained CLIP $F_{\mathrm{clip}}$, Batch size $B$, Learning rates $\mathrm{lr}_{sft},\mathrm{lr}_{rl}$, Epochs $\{E_{sft}, E_{rl}\}$, Confidence threshold $\tau$,  $\epsilon=2/255,\alpha=0.5/255, \text{ and }\mathrm{steps}=20$;}
\KwOut{Robust target model $F_{\mathrm{target}}$;}
\smallskip
\textbf{Stage 1: Supervised Fine-Tuning (SFT)}\\
 Initialize $F_{\mathrm{target}}\gets F_{\mathrm{clip}}$ with $\theta_{\mathrm{cls}}\sim\mathcal{N}(0,0.02^2)$\;
Freeze $\theta_{\mathrm{clip}} \setminus \{\mathrm{proj}\}$ (visual encoder)
\;
\For{$\mathrm{epoch}=1$ to $E_{sft}$}{
Sample $\{(x_s^{(i)},y_s^{(i)})\}_{i=1}^B \sim\mathcal{D}_s$\;
$x_{\mathrm{adv}}=\mathrm{PGD}(F_{\mathrm{target}},x_s,y_s; \epsilon,\alpha,\mathrm{steps})$\;
$L_{\mathrm{total}} = \underbrace{\mathrm{CE}(F_{\mathrm{target}}(x_s),y_s)}_{\text{Clean}} + \underbrace{\mathrm{CE}(F_{\mathrm{target}}(x_{\mathrm{adv}}),y_s)}_{\text{Adversarial}}$\;
$\theta_{\mathrm{cls}}\gets\theta_{\mathrm{cls}} - \mathrm{lr}_{sft}\nabla_{\theta}L_{\mathrm{total}}$\;
\If{$\mathrm{epoch}\ge2$} 
{
Unfreeze $\mathrm{proj}\in\theta_{\mathrm{clip}}$\; 
}
}
\medskip
\textbf{Stage 2: Reinforcement Adaptation (RA)}\\
$\tau\gets0.85$ (Initial confidence threshold)\;
\For{$\mathrm{cycle}=1$ to $E_{rl}$}
{
$\hat y_t=\arg\max_{y\in\mathcal{Y}} F_{\mathrm{target}}(x_t)\;\;\forall x_t\in\mathcal{D}_t$\;
$\mathcal{D}_{\mathrm{conf}}=\{(x_t,\hat y_t) \mid \max(\sigma(F_{\mathrm{target}}(x_t)))>\tau\}$\;
 $\mathcal{D}_{\mathrm{comb}} \gets \mathcal{D}_s \oplus \mathcal{D}_{\mathrm{conf}}$ (Balanced union)\;
\For{$\mathrm{epoch}=1$ to $2$}
{Sample $\{(x^{(i)},y^{(i)})\}_{i=1}^B \sim\mathcal{D}_{\mathrm{comb}}$
$x_{\mathrm{adv}}=\mathrm{PGD}(F_{\mathrm{target}},x,y; \epsilon,\alpha,\mathrm{steps})$\;
$L_{\mathrm{robust}} = \mathrm{CE}(F_{\mathrm{target}}(x),y) + \lambda\mathrm{CE}(F_{\mathrm{target}}(x_{\mathrm{adv}}),y)$\;
$\{\theta_{\mathrm{proj}},\theta_{\mathrm{cls}}\}\gets\{\theta_{\mathrm{proj}},\theta_{\mathrm{cls}}\} - \mathrm{lr}_{rl}\nabla_{\theta}L_{\mathrm{robust}}$\;
}
$\tau\gets\max(0.7,0.9\tau)$\;
}
\textbf{return} $F_{\mathrm{target}}$\; 
\end{algorithm}

In this work, we make two pivotal design decisions to guide the SFT process and enhance domain adaptability and robustness:

\noindent $\bullet$ \textbf{Partial Unfreezing Strategy:}
During the SFT phase, we adopt a partial unfreezing approach, wherein the parameters of the linear classifier $g(\cdot)$ and CLIP's final visual projection layer $\text{proj}(\cdot)$ are optimized, while the remaining layers of $f_{\text{CLIP}}(\cdot)$ are kept frozen. This selective fine-tuning allows the model to adapt to the source domain while retaining the generalization capabilities from CLIP pretraining.

\noindent $\bullet$ \textbf{Adversarial Robustness via PGD Training:}
To enhance robustness against adversarial perturbations, we employ adversarial training based on the PGD attack. The adversarial examples are generated on the fly using PGD and incorporated into the training objective during SFT. It promotes stability in model predictions under worst-case perturbations and enhances generalization across domains.

\subsection{Supervised Fine-Tuning Phase}
In the first stage of our training pipeline, we perform SFT on labelled source data $D_s = \{(x_i^s, y_i^s)\}_{i=1}^{N_s}$ 
while explicitly enforcing adversarial robustness. This phase adapts the classifier head $g(\cdot)$ and the CLIP visual projection layer $\text{Proj}(\cdot)$ to the source domain through a two-part procedure: (1) generating adversarial examples via PGD and (2) jointly optimizing on clean and perturbed inputs.

\subsubsection{Adversarial Example Generation}
For each mini-batch $(x_s, y_s)$, we generate adversarial perturbations $\delta$ by running a $T=20$-step PGD attack:

\paragraph{Initialization:}
\begin{equation}
\delta^{(0)} \sim \mathcal{U}(-\epsilon, \epsilon), \quad \epsilon = \frac{8}{255}.
\end{equation}

\paragraph{Iterative Update ($for$ $t = 0, \ldots, T-1$) at $\alpha = \frac{2}{255}$ :}
\begin{align}\small
\delta^{t+1} = \text{Clip}_{\|\delta\|_\infty \leq \epsilon} ( \delta^{t} + \alpha \cdot \text{sign} ( \nabla_{\delta} \mathcal{L}(h(x_s + \delta^{t}), y_s) ).
\end{align}

\paragraph{Adversarial Example:}
\begin{equation}
x_s^{\text{adv}} = x_s + \delta^{(T)}.
\end{equation}

The process ensures that each adversarial example $x_s^{\text{adv}}$ lies within an $\ell_\infty$-ball of radius $\epsilon$ around the original image, thus approximating worst-case perturbations under the given threat model.

\subsubsection{Robust Training Objective}
We form a combined batch after generating adversarial counterparts:
\begin{equation}
B_{\text{comb}} = \{(x_s, y_s)\} \cup \{(x_s^{\text{adv}}, y_s)\}.
\end{equation}

The fine-tuning loss is then defined as the average cross-entropy over clean and adversarial samples:

\begin{equation}\small
\mathcal{L}_{\text{SFT}} = \frac{1}{2|B_{\text{combined}}|} \sum_{(x, y) \in B_{\text{comb}}} \left[ -\sum_{c=1}^{C} y_c \log(\sigma(h(x))_c) \right],
\end{equation}
where $\sigma(\cdot)$ denotes the softmax function and $C$ is the classes count.

\paragraph{Learning Rate Schedule:}
To stabilize training, we adopt a two-phase learning rate schedule:

\begin{itemize}
    \item \textbf{Phase 1 (Epochs 1–2):} Only the classifier head $g(\cdot)$ is updated, using a learning rate $\eta_1 = 10^{-3}$.
    \item \textbf{Phase 2 (Epochs 3+):} We unfreeze the projection layer $\text{Proj}(\cdot)$ and jointly train $\{g, \text{Proj}\}$ with a reduced learning rate $\eta_2 = 10^{-5}$.
\end{itemize}

This curriculum allows the model to adapt its classifier weights to the source domain rapidly and then gently adjust the projection layer for improved feature alignment and robustness.


\subsection{Reinforcement Learning Phase}
In the second stage of our training framework, we adopt a reinforcement-inspired learning strategy rather than conventional reinforcement learning with explicit policy optimization. The design leverages feedback-driven updates and adaptive threshold decay, guided by the model's learning progress, to refine pseudo-labels and enhance adaptation. Specifically, both the classifier head $g(\cdot)$ and the projection layer $\mathrm{Proj}(\cdot)$ are progressively aligned with the target distribution $\mathcal{D}_t = \{x_i^t\}_{i=1}^{N_t}$. This phase consists of three core components: confidence-based pseudo-labeling, composite domain training with adversarial examples, and an adaptive confidence threshold schedule.


\subsubsection{Confidence-Based Pseudo-Labeling}
We first compute the model’s softmax output for each target image $x_t$: $p_t = \sigma(h(x_t)) \in \mathbb{R}^C$, where $C$ denotes the number of classes. We then assign a pseudo-label as: $\hat{y}_t = \arg\max_c p^c_t$ only if the maximum predicted probability exceeds a fixed threshold $\tau$, \textit{i.e.,} $\max_c p^c_t \geq \tau$. All samples failing this criterion are discarded for the current cycle, yielding a filtered pseudo-labeled set

\begin{equation}
    \mathcal{D}'_t = \{(x_i^t, \hat{y}_i^t) \mid \max_c p^{i,c}_t \geq \tau\}.
\end{equation}

\begin{table*}[htbp]
\centering
\caption{Comparison on PGD $20$ attack at $\epsilon = 2/255$ using PACS dataset. Avg accuracy calculated using tables (a) and (b).}
\label{tc3k}
\begin{tabular}{|l|l|l|l|l|l|l|l|l|l|l|l|l|}
\multicolumn{13}{c}{(a)}\\
\hline
$\textbf{S}\rightarrow \textbf{T}$ 
& \multicolumn{2}{c|}{$\textbf{Ph} \rightarrow \textbf{Ar}$} & 
\multicolumn{2}{c|}{$\textbf{Ph}\rightarrow\textbf{Ca}$} & 
\multicolumn{2}{c|}{$\textbf{Ph}\rightarrow\textbf{Sk}$} & 
\multicolumn{2}{c|}{$\textbf{Ca}\rightarrow\textbf{Ar}$} & 
\multicolumn{2}{c|}{$\textbf{Ca}\rightarrow\textbf{Ph}$}& 
\multicolumn{2}{c|}{$\textbf{Ca}\rightarrow\textbf{Sk}$}   \\ \hline

\textbf{Method} & Clean & PGD & Clean & PGD & Clean & PGD & Clean & PGD & Clean & PGD & Clean & PGD \\ \hline
DANN	
& $\mathbf{89.0}$ & $5.5$ &	 $80.5$	& $11.5$ &	 $74.2$ 	&	 $24.0$  & $84.8$ & $0.5$ &	 $92.3$ & $1.0$ &	$78.0$	&	 $25.8$\\ \hline
UDA+AT
& $82.1$ & $59.0$ &	 $85.2$	& $77.1$ &	 $78.0$ 	&	 $75.3$ & $76.2$ & $55.0$ &	 $93.1$ & $80.2$ &	$80.1$	&	 $77.2$\\ \hline
UDA + Trades 
& $82.0$ & $63.0$ &	 $84.2$	& $76.5$ &	 $78.6$ 	&	 $75.2$ & $78.5$ & $58.0$ &	 $92.2$ & $82.1$ &	$79.9$	&	 $77.6$\\ \hline
UDA + Mart 	
& $80.5$ & $62.7$ &	 $81.2$	& $77.6$ &	 $77.2$ 	&	 $75.8$ & $76.2$ & $57.8$ &	 $91.7$ & $82.6$ &	$79.1$	&	 $78.2$\\ \hline
ARTUDA 
& $85.9$ & $60.1$ &	 $87.5$	& $78.1$ &	 $74.9$ 	&	 $70.4$ & $76.5$ & $53.3$ &	 $89.4$ & $75.0$ &	$80.3$	&	 $74.9$\\ \hline
SRoUDA 
& $76.1$ & $56.4$ &	 $82.4$	& $71.7$ &	 $71.9$ 	&	 $63.7$ & $72.0$ & $50.9$ &	 $90.3$ & $79.9$ &	$76.7$	&	 $72.3$\\ \hline
DART 
& $85.2$ & $58.0$ &	 $\mathbf{89.4}$	& $\mathbf{80.5}$ &	 $\mathbf{82.5}$ 	&	 $\mathbf{79.9}$ & $77.4$ & $54.6$ &	 $94.2$ & $79.8$ &	$\mathbf{84.9}$	&	 $80.7$\\ \hline \hline
\textbf{Ours}
& $88.3$ & $\mathbf{75.4}$ &	 $78.8$	& $70.1$ &	 $78.2$ 	&	 $76.7$ & $\mathbf{94.6}$ & $\mathbf{79.6}$ &	 $\mathbf{95.5}$ & $\mathbf{87.9}$ &	$81.5$	&	 $\mathbf{83.5}$\\
\hline
\multicolumn{13}{c}{(b)}
\end{tabular}

\begin{tabular}{|l|l|l|l|l|l|l|l|l|l|l|l|l|l|l|}
\hline
$\textbf{S}\rightarrow \textbf{T}$ 
& \multicolumn{2}{c|}{$\textbf{Ar} \rightarrow \textbf{Ca}$} & 
\multicolumn{2}{c|}{$\textbf{Ar}\rightarrow\textbf{Ph}$} & 
\multicolumn{2}{c|}{$\textbf{Ar}\rightarrow\textbf{Sk}$} & 
\multicolumn{2}{c|}{$\textbf{Sk}\rightarrow\textbf{Ar}$} & 
\multicolumn{2}{c|}{$\textbf{Sk}\rightarrow\textbf{Ca}$}& 
\multicolumn{2}{c|}{$\textbf{Sk}\rightarrow\textbf{Ph}$} &
\multicolumn{2}{c|}{\textbf{Avg Accuracy}}   \\ \hline

\textbf{Method} & Clean & PGD & Clean & PGD & Clean & PGD & Clean & PGD & Clean & PGD & Clean & PGD & Clean & PGD\\ \hline
DANN	
& $84.2$ & $12.5$ &	 $97.9$	& $2.7$ &	 $84.9$ 	&	 $0.5$ 	& $68.0$ & $45$ &	 $72.1$ & $14.2$ &	$71.3$	&	 $0.1$ & $81.4$ & $11.9$
\\ \hline
UDA+AT
& $85.0$ & $76.3$ &	 $95.0$	& $83.1$ &	 $84.8$ 	&	 $81.1$ & $62.1$ & $37.5$ &	 $72.5$ & $64.0$ &	$88.8$	&	 $73.6$ & $81.9$ &$70.0$
 \\ \hline
UDA+Trades
& $85.1$ & $77.0$ &	 $98.3$	& $82.2$ &	 $86.0$ 	&	 $83.3$ & $69.0$ & $44.2$ &	 $76.6$ & $63.8$ &	$89.6$	&	 $76.4$ & $83.3$ & $71.6$
\\ \hline
UDA+Mart
& $84.9$ & $78.1$ &	 $95.5$	& $82.5$ &	 $85.2$ 	&	 $83.7$ & $67.5$ & $45.6$ &	 $71.8$ & $63.1$ &	$88.3$	&	 $\mathbf{77.1}$ & $81.6$ & $72.1$
 \\ \hline
ARTUDA 
& $88.1$ & $76.0$ &	 $95.0$	& $78.5$ &	 $80.3$ 	&	 $61.5$ & $49.5$ & $31.7$ &	 $38.1$ & $28.5$ &	$48.9$	&	 $40.4$ & $74.5$ & $60.7$
\\ \hline
SRoUDA
 & $84.2$ & $75.8$ &	 $94.1$	& $81.5$ &	 $77.3$ 	&	 $73.2$ & $24.8$ & $22.4$ &	 $72.4$ & $62.3$ &	$\mathbf{91.9}$	&	 $73.1$ & $76.2$& $65.3$
 \\ \hline
DART
& $89.1$ & $79.1$ &	 $98.15$	& $81.4$ &	 $\mathbf{89.5}$ 	&	 $\mathbf{86.4}$ & $71.9$ & $\mathbf{53.1}$ &	 $78.4$ & $69.2$ &	$87.8$	&	 $76.8$ & $85.7$ & $73.3$
 \\ \hline \hline
\textbf{Ours}
& $\mathbf{95.7}$ & $\mathbf{90.8}$ &	 $\mathbf{98.5}$	& $\mathbf{88.7}$ &	 $82.9$ 	&	 $83.1$ & $\mathbf{82.4}$ & $52.7$ &	 $\mathbf{89.7}$ & $\mathbf{80.3}$ &	$71.6$	&	 $59.3$ & $\mathbf{86.5}$ & $\mathbf{77.3}$\\ \hline
\end{tabular}
\end{table*}

\subsubsection{Composite Domain Training}
We merge the source data $\mathcal{D}_s$ with the high-confidence pseudo-labeled target set $\mathcal{D}'_t$ to form the composite training distribution.
\begin{equation}
    \mathcal{D}_{\mathrm{RL}} = \mathcal{D}_s \cup \mathcal{D}'_t.
\end{equation}

During each training iteration, we sample mini-batches $(x, y) \sim \mathcal{D}_{\mathrm{RL}}$ and generate adversarial counterparts $x_{\mathrm{adv}}$ via the same PGD procedure described in Section~2. The RL objective is then optimized over both clean and adversarial inputs:
\begin{equation}
    \mathcal{L}_{\mathrm{RL}} = \frac{1}{2}(\mathcal{L}(h(x), y) + \mathcal{L}(h(x_{\mathrm{adv}}), y)),
\end{equation}
where $\mathcal{L}(\cdot, \cdot)$ is the standard cross-entropy loss. This joint optimization encourages the model to maintain robustness while aligning representations across domains.

\subsubsection{Adaptive Confidence Threshold}
To progressively incorporate more target samples as the model’s predictions improve, we decay the pseudo-label confidence threshold $\tau$ after each RL cycle $k$:
\begin{equation}
    \tau^{(k+1)} = \max(0.7, 0.9 \tau^{(k)}).
\end{equation}
This schedule prevents the threshold from dropping below a minimum of $0.7$, ensuring that only sufficiently confident predictions contribute to training.

\subsubsection{Optimization Details}
Throughout the RL phase, we jointly update $\{g, \mathrm{Proj}\}$ using Adam with a fixed learning rate $\eta = 10^{-5}$. For each mini-batch, we perform two gradient steps: first minimizing $\mathcal{L}_{\mathrm{RL}}$ on adversarial examples, and then refining the pseudo-labeled set $\mathcal{D}'_t$ by re-computing confidences under the updated model. This alternating procedure continues for a pre-specified number of cycles or until convergence, resulting in a robust and well-adapted classifier to the target domain.

\section{Experiments}
This section evaluates the effectiveness and robustness of our proposed framework across multiple standard domain adaptation benchmarks, comparing against various competitive baselines. Our evaluation includes diverse datasets such as OfficeHome, PACS and VisDA, covering various domain shifts and classification challenges. We implement our method using CLIP’s ViT-B/32 backbone and generate adversarial examples using PGD-20 to test robustness. The experiments also include an ablation study of SFT+RL. For complete implementation details and reproducibility, please check SFT+RL Codebase\footnote{
https://l1nk.dev/t0pE8}.

\subsection{Datasets}
We conduct experiments on four standard multi-domain benchmarks. 

\noindent $\bullet$ \textbf{OfficeHome}~\cite{tomm-ude} comprises four distinct visual styles Art (Ar, 2,427 images), ClipArt (Cl, 4,365 images), Product (Pr, 4,439 images), and RealWorld (Re, 4,357 images) across 65 object categories. 

\noindent $\bullet$ \textbf{PACS}~\cite{pacs} contains four domains Photo (Ph, 1,670 images), Art Painting (Ar, 2,048 images), Cartoon (Ca, 2,344 images), and Sketch (Sk, 3,929 images) spanning 7 classes. 

\noindent $\bullet$ \textbf{VisDA}~\cite{visda} is a large synthetic to real adaptation benchmark with 12 classes, including 152,409 synthetic and 55,400 real images.

\subsection{Baselines}
We compare against seven competitive baselines:
\begin{itemize}
    \item \textbf{Vanilla UDA}: A standard UDA model trained without robustness constraints.
    \item \textbf{UDA + AT}: Uses a DANN feature-alignment model~\cite{dann} to generate pseudo-labels on the target data followed by adversarial training.
    \item \textbf{UDA + TRADES}: Substitutes the TRADES objective~\cite{zhang2019theoretically} for adversarial loss.
    \item \textbf{UDA + MART}: Replaces the adversarial loss with MART~\cite{madry2017towards}.
    \item \textbf{ARTUDA}~\cite{yang2021exploring}: A recent state-of-the-art adversarial UDA approach.
    \item \textbf{SRoUDA}~\cite{zhu2023srouda}: Self-training with robust pseudo-labels.
    \item \textbf{DART}~\cite{wang2024dart}: Domain-aware adversarial robustness framework.
\end{itemize}

\begin{table*}[htbp]
\centering
\caption{Comparison on PGD $20$ attack at $\epsilon = 2/255$ using OfficeHome. Avg accuracy calculated using tables (a) and (b).}
\label{tc2}
\begin{tabular}{|l|l|l|l|l|l|l|l|l|l|l|l|l|}
\multicolumn{13}{c}{(a)}\\
\hline
$\textbf{S}\rightarrow \textbf{T}$ 
& \multicolumn{2}{c|}{$\textbf{Pr} \rightarrow \textbf{Ar}$} & 
\multicolumn{2}{c|}{$\textbf{Pr}\rightarrow\textbf{Cl}$} & 
\multicolumn{2}{c|}{$\textbf{Pr}\rightarrow\textbf{Re}$} & 
\multicolumn{2}{c|}{$\textbf{Cl}\rightarrow\textbf{Ar}$} & 
\multicolumn{2}{c|}{$\textbf{Cl}\rightarrow\textbf{Pr}$}& 
\multicolumn{2}{c|}{$\textbf{Cl}\rightarrow\textbf{Re}$}   \\ \hline
\textbf{Method} $\downarrow$ & Clean & PGD & Clean & PGD & Clean & PGD & Clean & PGD & Clean & PGD & Clean & PGD \\ \hline
DANN 	
& $49.1$ & $0.2$ &	 $\mathbf{57.4}$	& $2.1$ &	 $60.0$ 	&	 $0.2$ 	& $45.2$ & $0.0$ &	 $47.9$ & $3.6$ &	$67.4$	&	 $0.6$\\ \hline
UDA+AT
& $38.5$ & $18.3$ &	 $49.1$	& $42.9$ &	 $61.4$ 	&	 $44.2$  & $39.4$ & $23.0$ &	 $55.6$ & $46.8$ &	$56.5$	&	 $41.5$\\ \hline
UDA+Trades
& $37.9$ & $20.5$ &	 $49.3$	& $44.3$ &	 $61.6$ 	&	 $44.0$ & $40.0$ & $22.0$ &	 $56.2$ & $47.9$ &	$56.2$	&	 $43.8$\\ \hline
UDA+Mart 	
& $37.6$ & $26.0$ &	 $48.1$	& $46.0$ &	 $61.8$ 	&	 $44.8$ & $38.5$ & $24.5$ &	 $55.8$ & $48.1$ &	$56.8$	&	 $44.6$\\ \hline
ARTUDA 
& $38.3$ & $18.0$ &	 $48.5$	& $42.8$ &	 $62.4$ 	&	 $42.7$ & $42.0$ & $20.2$ &	 $56.1$ & $44.1$ &	$58.9$	&	 $30.2$\\ \hline
SRoUDA
& $33.5$ & $22.4$ &	 $49.9$	& $41.6$ &	 $60.2$ 	&	 $45.6$ & $36.3$ & $23.8$ &	 $53.9$ & $47.2$ &	$55.1$	&	 $42.1$\\ \hline
DART 
& $43.7$ & $21.5$ &	 $52.3$	& $44.8$ &	 $63.5$ 	&	 $43.6$ & $44.1$ & $24.2$ &	 $57.0$ & $45.5$ &	$57.8$	&	 $39.6$\\ \hline \hline
\textbf{Ours}
& $\mathbf{62.1}$ & $\mathbf{41.7}$ &	 $52.5$	& $\mathbf{46.2}$ &	 $\mathbf{76.5}$ 	&	 $\mathbf{55.6}$ & $\mathbf{70.4}$ & $\mathbf{43.6}$ &	 $\mathbf{70.4}$ & $\mathbf{58.8}$ &	$\mathbf{78.6}$	&	 $\mathbf{60.4}$\\ \hline
\multicolumn{13}{c}{(b)}
\end{tabular}

\begin{tabular}{|l|l|l|l|l|l|l|l|l|l|l|l|l|l|l|}
\hline
$\textbf{S}\rightarrow \textbf{T}$ 
& \multicolumn{2}{c|}{$\textbf{Re} \rightarrow \textbf{Ar}$} & 
\multicolumn{2}{c|}{$\textbf{Re}\rightarrow\textbf{Cl}$} & 
\multicolumn{2}{c|}{$\textbf{Re}\rightarrow\textbf{Pr}$} & 
\multicolumn{2}{c|}{$\textbf{Ar}\rightarrow\textbf{Cl}$} & 
\multicolumn{2}{c|}{$\textbf{Ar}\rightarrow\textbf{Pr}$}& 
\multicolumn{2}{c|}{$\textbf{Ar}\rightarrow\textbf{Re}$}  & 
\multicolumn{2}{c|}{\textbf{Avg Accuracy}}   \\ \hline \hline
\textbf{Method} & Clean & PGD & Clean & PGD & Clean & PGD & Clean & PGD & Clean & PGD & Clean & PGD & Clean & PGD \\ \hline
DANN	
& $66.0$ & $0.4$ &	 $55.5$	& $3.6$ &	 $74.3$ 	&	 $1.8$ 	& $49.1$ & $2.6$ &	 $55.5$ & $0.9$ &	$66.8$	&	 $1.0$ & $57.8$ & $1.4$  \\ \hline
UDA+AT
& $46.4$ & $29.4$ &	 $55.0$	& $49.4$ &	 $72.3$ 	&	 $60.4$  & $48.0$ & $41.7$ &	 $55.9$ & $46.2$ &	$57.6$	&	 $40.5$& $52.9$ & $40.3$ \\ \hline
UDA+Trades 
& $47.9$ & $27.4$ &	 $55.6$	& $49.7$ &	 $70.0$ 	&	 $61.9$ & $48.6$ & $43.6$ &	 $55.9$ & $47.6$ &	$57.1$	&	 $41.8$& $53.0$ & $41.2$  \\ \hline
UDA+Mart 	
& $46.8$ & $28.7$ &	 $56.1$	& $\mathbf{50.6}$ &	 $71.2$ 	&	 $62.5$ & $48.3$ & $43.9$ &	 $54.8$ & $47.9$ &	$58.3$	&	 $42.6$& $52.8$ & $42.1$  \\ \hline
ARTUDA~\ 
& $49.5$ & $28.4$ &	 $58.3$	& $48.5$ &	 $73.0$ 	&	 $58.3$ & $50.9$ & $41.7$ &	 $55.0$ & $41.2$ &	$61.7$	&	 $42.5$ & $54.6$ & $38.2$ \\ \hline
SRoUDA
 & $42.1$ & $27.5$ &	 $55.4$	& $47.3$ &	 $70.7$ 	&	 $60.6$ & $48.2$ & $38.9$ &	 $52.9$ & $45.8$ &	$57.4$	&	 $44.2$ & $51.3$ & $40.5$ \\ \hline
DART 
& $53.7$ & $29.1$ &	 $\mathbf{58.6}$	& $49.8$ &	 $74.0$ 	&	 $60.2$ & $50.4$ & $42.2$ &	 $60.1$ & $47.7$ &	$62.7$	&	 $40.7$& $56.5$ & $40.7$\\ \hline \hline
\textbf{Ours}
& $\mathbf{70.4}$ & $\mathbf{50.0}$ &	 $56.5$	& $49.1$ &	 $\mathbf{78.6}$ 	&	 $\mathbf{67.1}$ & $\mathbf{63.3}$ & $\mathbf{53.9}$ &	 $\mathbf{73.5}$ & $\mathbf{64.5}$ &	$\mathbf{79.5}$	&	 $\mathbf{63.4}$ & $\mathbf{69.4}$ & $\mathbf{54.5}$\\ \hline
\end{tabular}
\end{table*}

\subsection{Implementation Details}
Our implementation builds upon CLIP’s ViT-B/32 image encoder~\cite{radford2021learning} as a frozen backbone during the initial supervised fine-tuning (SFT) stage, with a linear classifier head mapping the 512-dimensional feature space to the $C$ target classes. 

Adversarial examples are generated on-the-fly via a 20-step PGD attack 
~\cite{madry2017towards}. Optimization uses Adam with two learning-rate schedules: 
\begin{itemize}
    \item During SFT: Classifier and CLIP projection layer use learning rates of $10^{-3}$ and $10^{-5}$, respectively.
    \item During RL: Both Classifier and CLIP projection layer use $10^{-5}$.
\end{itemize}

Due to memory constraints from adversarial generation, we set the batch size to 8. The SFT phase runs for 10 epochs (with the projection layer unfrozen after epoch 2), followed by three RL cycles of two epochs each, starting from a pseudo-label confidence threshold $\tau=0.85$ decayed by 0.9 per cycle. 

All images are resized to $224\times224$, center-cropped, and normalized using CLIP’s statistics. During both phases, we jointly minimize cross-entropy over clean and adversarial batches, and report clean and robust accuracy (under PGD-20) averaged over three runs.

\subsection{Results}

\subsubsection{Comparison under PGD-20 Attack ($\epsilon = 2/255$)}

Tables~\ref{tc3k} and~\ref{tc2} report the robust accuracy of various methods on the PACS and OfficeHome datasets, respectively, when subjected to a white box PGD attack with 20 steps and perturbation magnitude $\epsilon = 2/255$. In all experiments, DANN serves as the UDA baseline.  

\begin{figure}[h]
\centering
\includegraphics[width=\columnwidth]{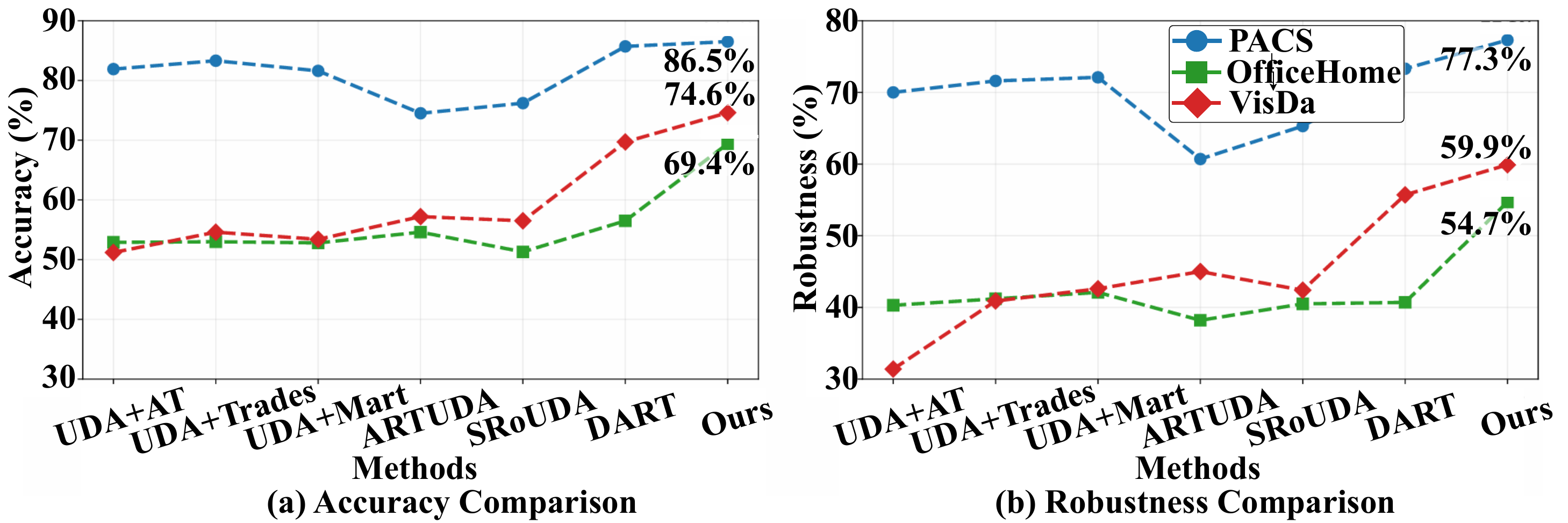}
\caption{Comparison of robustness and accuracy across the OfficeHome, VisDA, and PACS datasets.}
\vspace{0.5cm}
\label{fig:accuracy_robustness}
\end{figure}

From the results, our proposed \textbf{SFT+RL} method consistently outperforms both conventional UDA baselines and recent state of the art adversarial UDA approaches in terms of both adversarial robustness and clean accuracy. Notably, on the OfficeHome dataset, \textbf{SFT+RL} improves average robustness from 1.4\% to 54.5\% and enhances clean accuracy from 57.8\% to 69.5\% compared to the UDA baseline DANN. Similarly, on the PACS dataset, it boosts robustness from 11.9\% to 77.3\% and increases clean accuracy from 81.4\% to 86.5\% over DANN. These performance trends are illustrated across all domains in Figure~\ref{fig:accuracy_robustness}.

\subsubsection{Evaluation under PGD-20 Attack ($\epsilon = 8/255$)}

Table~\ref{tc4} compares the robustness and clean accuracy of various methods on the VisDA dataset under a white box PGD attack with 20 steps and perturbation magnitude $\epsilon = 8/255$, using DANN as the UDA baseline. Our proposed \textbf{SFT+RL} method achieves substantial gains, with average adversarial robustness increasing from 0.4\% up to 55.7\%, and average clean accuracy rising from 73.0\% to 74.6\% over competing approaches.

\begin{figure}[h]
\centering
\includegraphics[width=7.5cm]{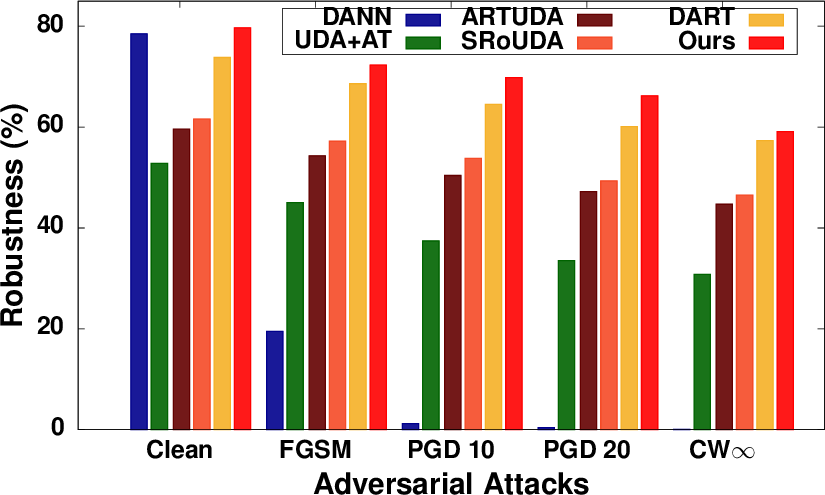}
\caption{Robustness comparison against different adversarial
attacks (FGSM, PGD 10, PGD 20, CW$\infty$) for various methods using VISDA dataset (Real$\rightarrow$synthetic).}
\label{k5}
\vspace{0.35cm}
\end{figure}

In addition, we evaluate robustness on the Real~$\rightarrow$~Synthetic adaptation pair of VisDA across multiple attack types (FGSM, PGD-10, PGD-20, and CW$_\infty$). The results, reported in Figure~\ref{k5}, demonstrate that \textbf{SFT+RL} consistently outperforms standard UDA baselines and recent adversarial UDA methods under diverse threat models. \vspace{0.4cm}

\begin{table}[htbp]
\centering
\caption{An illustration of comparison on PGD $20$ attack at $\epsilon = 8/255$ using VisDA dataset.}
\label{tc4}
\begin{tabular}{|l|ll|ll|ll|}
\hline
$\textbf{S} \rightarrow \textbf{T} $ & \multicolumn{2}{c|}{$ \textbf{Syn} \rightarrow \textbf{Re}$} & \multicolumn{2}{c|}{$\textbf{Re}\rightarrow \textbf{Syn}$} & \multicolumn{2}{c|}{\textbf{Avg Accuracy}} \\ \hline
~\textbf{Method}~ & \multicolumn{1}{l|}{Clean} & PGD & \multicolumn{1}{l|}{Clean} & PGD & \multicolumn{1}{l|}{Clean} & PGD \\ \hline
DANN~ & \multicolumn{1}{l|}{$67.5$} & $0.3$ & \multicolumn{1}{l|}{$78.5$} & $0.5$ & \multicolumn{1}{l|}{$73.0$} & $0.4$ \\ \hline
UDA+AT~ & \multicolumn{1}{l|}{$49.6$} & $29.3$ & \multicolumn{1}{l|}{$52.8$} & $33.5$ & \multicolumn{1}{l|}{$51.2$} & $31.4$ \\ \hline
Trades~ & \multicolumn{1}{l|}{$51.7$} & $36.1$ & \multicolumn{1}{l|}{$57.4$} & $45.6$ & \multicolumn{1}{l|}{$54.6$} & $40.9$ \\ \hline
Mart~ & \multicolumn{1}{l|}{$50.8$} & $38.4$ & \multicolumn{1}{l|}{$ 56.0$} & $46.8$ & \multicolumn{1}{l|}{$53.4$} & $42.6$\\ \hline
ARTUDA~& \multicolumn{1}{l|}{$54.9$} & $42.7$ & \multicolumn{1}{l|}{$59.5$} & $47.2$ & \multicolumn{1}{l|}{$57.2$} & $45.0$ \\ \hline
SRoUDA~& \multicolumn{1}{l|}{$51.3$} & $35.4$ & \multicolumn{1}{l|}{$61.6$} & $49.3$ & \multicolumn{1}{l|}{$56.5$} & $42.4$ \\ \hline
DART~& \multicolumn{1}{l|}{$65.6$} & $51.2$ & \multicolumn{1}{l|}{$73.8$} & $60.1$ & \multicolumn{1}{l|}{$69.7$} & $55.7$\\ \hline \hline
~\textbf{Ours}  ~& \multicolumn{1}{l|}{$\mathbf{69.5}$} & $\mathbf{53.6}$ & \multicolumn{1}{l|}{$\mathbf{79.7}$} & $\textbf{66.2}$ & \multicolumn{1}{l|}{$\mathbf{74.6}$} & $\mathbf{59.9}$\\ \hline
\end{tabular}
\end{table}

\subsection{Component Ablation of SFT+RL}
To evaluate the individual contributions of the Supervised Fine-Tuning (SFT) and Reinforcement Learning (RL) stages, we conduct ablation experiments on the OfficeHome dataset using the Art\,$\rightarrow$\,Clipart and Art\,$\rightarrow$\,Product transfer tasks under a PGD attack with $\epsilon = 2/255$. Table~\ref{a6} presents clean and adversarial (PGD) accuracies for the following configurations:

\begin{enumerate}[nosep]
  \item \textbf{DANN}: A standard UDA baseline without adversarial training,
  \item \textbf{UDA + AT}: Adversarial training applied to DANN,
  \item \textbf{SFT w/o RL}: Only the SFT stage is applied (no RL),
  \item \textbf{RL w/o SFT}: Only the RL stage is applied (no SFT),
  \item \textbf{SFT+RL}: Full two-stage method combining SFT and RL.
\end{enumerate}

Our observations are as follows:
\begin{itemize}[nosep]
  \item \emph{SFT w/o RL} improves performance, achieving gains of 9.4\% in clean accuracy and 11.4\% in robustness over UDA + AT baseline.
  \item \emph{RL w/o SFT} enhances the results, yielding improvements of 8.9\% in clean accuracy and 10.2\% in robustness over SFT w/o RL.
  \item The combined \textbf{SFT+RL} achieves the best results, with a clean accuracy of 68.4\% and PGD robustness of 59.2\%, reflecting gains of 10.5\% in clean accuracy and 9.8\% in robustness compared to RL w/o SFT.\\
\end{itemize}

\begin{table}[htbp]
\centering
\caption{Results on component ablations of SFT+RL at perturbation size $\epsilon = 2/255$ on OfficeHome dataset.}
\label{a6}
\begin{tabular}{|l|ll|ll|ll|}
\hline
\textbf{S$\rightarrow$T} & \multicolumn{2}{c|}{\textbf{Art$\rightarrow$Clipart}} & \multicolumn{2}{c|}{\textbf{Art$\rightarrow$Product}} & \multicolumn{2}{c|}{\textbf{Average Accuracy}} \\ \hline
\textbf{Method} & \multicolumn{1}{l|}{\textbf{Clean}} & \textbf{PGD} & \multicolumn{1}{l|}{\textbf{Clean}} & \textbf{PGD} & \multicolumn{1}{l|}{\textbf{Clean}} & \textbf{PGD} \\ \hline
DANN & \multicolumn{1}{l|}{$49.1$} & $2.6$ & \multicolumn{1}{l|}{$55.5$} & $0.9$ & \multicolumn{1}{l|}{$52.3$} & $1.7$ \\ \hline
UDA + AT & \multicolumn{1}{l|}{$48.0$} & $41.7$ & \multicolumn{1}{l|}{$55.9$} & $46.2$ & \multicolumn{1}{l|}{$51.9$} & $43.9$ \\ \hline
SFT w/o RL & \multicolumn{1}{l|}{$52.3$} & $45.6$ & \multicolumn{1}{l|}{$61.4$} & $52.2$ & \multicolumn{1}{l|}{$56.8$} & $48.9$ \\ \hline
RL w/o SFT & \multicolumn{1}{l|}{$56.8$} & $49.2$ & \multicolumn{1}{l|}{$67.1$} & $58.7$ & \multicolumn{1}{l|}{$61.9$} & $53.9$ \\ \hline
\textbf{SFT+RL} & \multicolumn{1}{l|}{$\textbf{63.3}$} & $\textbf{53.9}$ & \multicolumn{1}{l|}{$\textbf{73.5}$} & $\textbf{64.5}$ & \multicolumn{1}{l|}{$\textbf{68.4}$} & $\textbf{59.2}$ \\ \hline
\end{tabular}
\end{table}

\begin{figure}[h]
\centering
\includegraphics[width=\columnwidth]{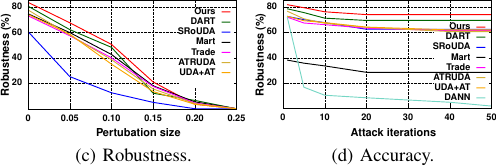}
\caption{Robustness as a function of perturbation size for various methods on officeHome (Art$\rightarrow$Clipart).}
\vspace{0.5cm}
\label{fig:perturbation_robustness}
\end{figure}

\subsubsection{Evaluation on Attack Budget}
We assess the scalability and robustness of our proposed \textbf{SFT+RL} method under varying adversarial attack budgets by evaluating two factors: (1) increasing the perturbation size $\epsilon$, and (2) increasing the number of PGD attack iterations while keeping $\epsilon = 2/255$. 

As illustrated in Figure~\ref{fig:perturbation_robustness} (c), increasing the perturbation size negatively impacts robustness accuracy across all methods. These evaluations are conducted on the PACS dataset for the Art$\rightarrow$Clipart transfer setting. While all models exhibit a decline in robustness as the perturbation size increases, our method consistently outperforms other approaches across all levels of perturbation. Notably, when the perturbation size becomes sufficiently large, all methods eventually degrade to near-zero robustness.

Furthermore, Figure~\ref{fig:perturbation_robustness} (d) shows that increasing the number of PGD iterations improves attack strength. In particular, robustness performance saturates at around 20 iterations, suggesting that PGD with 20 steps provides a sufficiently strong adversarial attack under a fixed perturbation size of $\epsilon = 2/255$.

\subsection{Sensitivity Analysis of Confidence Threshold and Decay}
\label{sec:threshold_sensitivity}

To evaluate the impact of pseudo-label confidence filtering, we conducted a sensitivity analysis by varying the confidence threshold and incorporating a decay strategy. The confidence threshold directly influences the quality and quantity of pseudo-labels selected during training: overly strict thresholds reduce the number of reliable pseudo-labels and hinder adaptation, while excessively relaxed thresholds introduce noisy labels that compromise robustness. Introducing a decay mechanism stabilizes training by gradually relaxing the filtering criterion, thereby allowing more pseudo-labels to be leveraged as the model adapts.

\begin{table}[ht]
\centering
\caption{Sensitivity analysis of confidence threshold and its decay on clean and adversarial target domain accuracy.}
\label{tab:threshold_sensitivity}
\begin{tabular}{|c|c|c|c|}
\hline
\textbf{Study} & \textbf{Confidence Threshold} & \textbf{Clean Acc. (\%)} & \textbf{Adv. Acc. (\%)} \\ \hline
1 & 0.85 (with decay) & \textbf{63.30} & \textbf{53.90} \\
2 & 0.70 & 60.60 & 53.49 \\
3 & 0.75 & 52.92 & 44.22 \\
4 & 0.80 & 56.01 & 48.57 \\
5 & 0.85 & 60.60 & 53.15 \\ \hline
\end{tabular}
\end{table}

As shown in Table~\ref{tab:threshold_sensitivity}, the adaptive decay setting (Study~1) achieved the best trade-off, with the highest clean target domain accuracy (63.3\%) while also preserving adversarial robustness (53.9\%). In contrast, fixed thresholds (Studies~3 and~4) resulted in significant drops in both clean and adversarial performance, highlighting the sensitivity of adaptation to threshold selection. These findings emphasize the importance of jointly tuning the confidence threshold and its decay to achieve stable adaptation and improved robustness across domains.

\section{Conclusion and Future Work}
In this work, we introduced SFT+RL, a novel two-phase framework for source-free adversarial adaptation of vision-language models such as CLIP. Our method combines adversarial fine-tuning with confidence-guided pseudo-labeling to effectively mitigate adversarial domain shifts and maintain cross-modal alignment. Through empirical evaluation, SFT+RL demonstrates substantial gains in adversarial robustness and clean accuracy compared to the baseline CLIP model. Key contributions include partial CLIP unfreezing to balance stability and generalization, a confidence-based curriculum for refined pseudo-labeling, and a cross-modal cohesion loss to preserve visual-textual alignment under adversarial conditions.

In future, we will expand the framework to accommodate larger multimodal models and explore RL-based strategies for adaptive threshold decay. We also plan to integrate parameter-efficient tuning methods such as LoRA and adapters, which can enhance adaptability while minimizing computation. These directions will further strengthen the applicability of SFT+RL in real-world, safety-critical environments, building on its demonstrated robustness.

\bibliography{references}

\end{document}